\documentclass[runningheads]{llncs}
\usepackage{graphicx}
\usepackage{multirow}
\usepackage{todonotes}
\usepackage{xcolor, soul}
\usepackage{comment}
\usepackage{adjustbox}
\usepackage{amsmath}
\usepackage[utf8]{inputenc}
\usepackage[english]{babel}
\usepackage{color}
\usepackage{amsfonts}
\usepackage{url}
\usepackage[mathscr]{euscript}
 \let\mathscr\relax% just so we can load this and rsfs
\usepackage[scr]{rsfso}
\usepackage{caption}
\usepackage[labelfont=bf]{caption}
\usepackage{tabularx,booktabs}
\newcolumntype{C}{>{\centering\arraybackslash}X} % centered version of "X" type
\usepackage{multicol}
\usepackage{algorithm2e}
\usepackage{booktabs,ragged2e}
\usepackage[utf8]{inputenc}
\usepackage{todonotes}
\usepackage[misc,geometry]{ifsym}
\usepackage{lineno}
\usepackage{makecell}

\usepackage[flushleft]{threeparttable}
\usepackage{subcaption}
\usepackage{color}

\author{Rebecca S. Stone(\Letter), Pedro E. Chavarrias-Solano, Andrew J. Bulpitt, David C. Hogg, Sharib Ali(\Letter) \orcidID{0000-0003-1313-3542}}
\authorrunning{R.S. Stone, P.E.C. Solano, A.J. Bulpitt, D.C. Hogg, S. Ali}
\institute{School of Computing, University of Leeds, LS2 9JT, Leeds, UK \\ sc16rsmy, s.s.ali@leeds.ac.uk}
\begin{document}

\title{Bayesian uncertainty-weighted loss for improved generalisability on polyp segmentation task}
\titlerunning{Bayesian uncertainty-weighted loss for polyp segmentation}
% \author{*}
% \institute{*}
\maketitle      
\begin{abstract}
While several previous studies have devised methods for segmentation of polyps, most of these methods are not rigorously assessed on multi-center datasets. Variability due to appearance of polyps from one center to another, difference in endoscopic instrument grades, and acquisition quality result in methods with good performance on in-distribution test data, and poor performance on out-of-distribution or underrepresented samples. Unfair models have serious implications and pose a critical challenge to clinical applications. We adapt an implicit bias mitigation method which leverages Bayesian predictive uncertainties during training to encourage the model to focus on underrepresented sample regions. We demonstrate the potential of this approach to improve generalisability without sacrificing state-of-the-art performance on a challenging multi-center polyp segmentation dataset (PolypGen) with different centers and image modalities.
%~\cite{ali2023multi}

% \keywords{Colonoscopy\and Inflammation \and Self-supervised learning  \and Classification \and Group discrimination \and Colitis}
\end{abstract}
% %
\section{Introduction}
% Clinical motivation and context - Sharib
% Brief on bias and what is motivation from medical perspective - Sharib
% Brief on why this paper - Rebecca 
Colorectal cancer (CRC) is the third most common cancer worldwide~\cite{silva2014toward} with early screening and removal of precancerous lesions (colorectal adenomas such as ``polyps'') suggesting longer survival rates. While surgical removal of polyps (polypectomy) is a standard procedure during colonoscopy, detecting these and their precise delineation, especially for sessile serrated adenomas/polyps, is extremely challenging. Over a decade, advanced computer-aided methods have been developed and most recently machine learning (ML) methods have been widely developed by several groups. However, the translation of these technologies in clinical settings has still not been fully achieved. One of the main reasons is the generalisability issues with the ML methods~\cite{ali2022}. Most techniques are built and adapted over carefully curated datasets which may not match the natural occurrences of the scene during colonoscopy.

% Motivate unfairness as a critical problem in medical imaging...
Recent literature demonstrates how intelligent models can be systematically unfair and biased against certain subgroups of populations. In medical imaging, the problem is prevalent across various image modalities and target tasks; for example, models trained for lung disease prediction~\cite{seyyed2020chexclusion}, retinal diagnosis~\cite{burlina2021addressing}, cardiac MR segmentation~\cite{puyol2021fairness}, and skin lesion detection~\cite{abbasi2020risk,li2021estimating} are all subject to biased performance against one or a combination of underrepresented gender, age, socio-economic, and ethnic subgroups. Even under the assumption of an ideal sampling environment, a perfectly balanced dataset does not ensure unbiased performance as relative quantities are not solely responsible for bias~\cite{wang2020mitigating,mehrabi2021survey}. This, and the scarcity of literature exploring bias mitigation for polyp segmentation in particular, strongly motivate the need for development and evaluation of mitigation methods which work on naturally occurring diverse colonoscopy datasets such as PolypGen~\cite{ali2023multi}.
\section{Related work}
% Medical image segmenation in focus to colonoscopy -  Peter
% Bias related methods in general (Rebecca)
% and then in medical imaging (and Sharib)

% TODO: read this paper and motivate -> OOD --> \textbf{https://arxiv.org/pdf/2306.07792.pdf} - 
% https://link.springer.com/book/10.1007/978-3-031-16749-2 => UNSURE Miccai workshophttps://openaccess.thecvf.com/content/CVPR2021/papers/Zhang_Distribution_Alignment_A_Unified_Framework_for_Long-Tail_Visual_Recognition_CVPR_2021_paper.pdf - class imbalance in long tail distribution
% OOD CVPR: https://openaccess.thecvf.com/content/CVPR2022/papers/Cao_Deep_Hybrid_Models_for_Out-of-Distribution_Detection_CVPR_2022_paper.pdf
% TMI: https://ieeexplore.ieee.org/stamp/stamp.jsp?tp=&arnumber=10144095
% 

% Polyp segmentation algorithms can be classified into two main categories: traditional and deep learning-based. Recent studies have shown that traditional methods which rely on low-level features have a worst performance when compared to deep learning-based approaches \cite{guo2022}. Recent deep learning-based methods for polyp segmentation leverage the use of convolutional neural networks (CNN) following an encoder-decoder architecture, such as, U-Net \cite{ronneberger2015}. Yue \textit{et al.} \cite{yue2023}.

% INTRODUCTION TO THE TASK (polyp segmentation) and the problem
Convolutional neural networks have recently worked favourably towards the advancement of building data-driven approaches to polyp segmentation using deep learning. These methods~\cite{mahmud2021,yeung2021} are widely adapted from the encoder-decoder U-Net~\cite{ronneberger2015} architecture. Moreover, addressing the problem of different polyp sizes using multi-scale feature pruning methods, such as atrous-spatial pyramid pooling in DeepLabV3~\cite{chen2018} or high-resolution feature fusion networks like HRNet~\cite{simonyan2014} have been used by several groups for improved polyp segmentation. For example, MSRFNet~\cite{Abhishek22} uses feature fusion networks between different resolution stages.
% THE PROBLEM: generalisability on out-of-distribution and minority/underrepresented populations
Recent work on generalisability assessment found that methods trained on specific centers do not tend to generalise well on unseen center data or different naturally occurring modalities such as sequence colonoscopy data~\cite{ali2022}. These performance gaps were reported to be large (drops of nearly 20\%).

% LIT: the problem and literature in the CV community in general \cite{shen2021towards}
Out-of-distribution (OOD) generalisation and bias mitigation are challenging, open problems in the computer vision research community. While in the bias problem formulation, models wrongly correlate one or more spurious (non-core) features with the target task, the out-of-distribution problem states that test data is drawn from a separate distribution than the training data. Some degree of overlap between the two distributions in the latter formulation exists, which likely includes the core features. Regardless of the perspective, the two problems have clear similarities, and both result in unfair models which struggle to generalise for certain sub-populations.
% OOD in literature; CV and polyp segmentation
In the literature, many works focus on OOD detection, through normal or modified softmax outputs~\cite{hendrycks2019using}, sample uncertainty thresholds from Bayesian, ensemble, or other models~\cite{mehrtash2020confidence,jungo2020analyzing,cao2022deep}, and distance measures in feature latent space~\cite{gonzalez2021detecting}. Other approaches tackle the more difficult problem of algorithmic mitigation through disentangled representation learning, architectural and learning methods, and methods which optimise for OOD generalisability directly~\cite{shen2021towards}.
% aligning features in the latent space~\cite{ji2023rethinking}.

Similarly, several categories of bias mitigation methods exist. Some methods rely on two or more models, one encouraged to learn the biased correlations of the majority, and the other penalised for learning the correlations of the first~\cite{nam2020learning,kim2022learning}. Other approaches modify the objective loss functions to reward learning core rather than spurious features~\cite{xu2021consistent,pezeshki2021gradient}, or by neutralising representations to remove learned spurious correlations~\cite{du2021fairness}. Others use data augmentation~\cite{burlina2021addressing}, or explore implicit versions of up-weighting or re-sampling underrepresented samples by discovering sparse areas of the feature space~\cite{amini2019uncovering} or dynamically identifying samples more likely to be underrepresented~\cite{stone2022epistemic}. De-biasing methods leveraging Bayesian model uncertainties~\cite{khan2019striking,branchaud2021can,stone2022epistemic} provide the added benefits of uncertainty estimations which are useful in clinical application for model interpretability and building user confidence.

To tackle the generalisability problem for polyp segmentation, we consider the diversity of features in a multi-centre polyp dataset~\cite{ali2023multi}. Our contributions can be listed as:  1) adapting an implicit bias mitigation strategy in~\cite{stone2022epistemic} from a classification to a segmentation task; 2) evaluating the suitability of this approach on three separate test sets which have been shown to be challenging generalisation problems. 
Our experiments demonstrate that our method is comparable and in many cases even improves the performance compared to the baseline state-of-the-art segmentation method while decreasing performance discrepancies between different test splits.
% It should be anonymous so we cannot write straightforward - lets rewrite it if it gets accepted! - OK got it. R
%
%
% We make the following contributions: 1) we modify the implicit bias mitigation introduced in~\cite{stone2022epistemic} from 
% a classification to a segmentation task and apply it to the challenging multi-center polyp detection and segmentation problem; and 2) we evaluate the suitability of this approach on three separate test datasets for bias mitigation and out-of-distribution generalisation, showing that our method is comparable or better than state-of-the-art performance while decreasing performance discrepancies between the test datasets.

% comment on Bayesian epistemic uncertainty and how it relates to bias - Rebecca
\section{Method}
% Rebecca and Peter
The encoder-decoder architecture for semantic segmentation has been widely explored in medical image analysis. In our approach we have used DeepLabV3~\cite{chen2018encoder} as baseline model that has SOTA performance on the PolypGen dataset~\cite{ali2023multi}. 
% We use a DeepLabV3~\cite{chen2018encoder} for our base architecture which has state-of-the-art performance on the PolypGen dataset~\cite{ali2023multi}. DeepLabV3+ builds on the semantic segmentation architecture introduced in DeepLabV3~\cite{chen2017rethinking} by adding a decoder which recovers spatial information.
We then apply a probabilistic model assuming a Gaussian prior on all trainable weights (both encoder and decoder) that are updated to the posterior using the training dataset. For the Bayesian network with parameters $\boldsymbol{\theta}$, posterior $p(\boldsymbol{\theta}~\mid D)$, training data with ground truth segmentation masks \(D = (X, Y)\), and sample $x_i$, the predictive posterior distribution for a given ground truth segmentation mask $y_i$ can be written as:
% To convert the DeepLabV3+ into a probabilistic model, we assume a Gaussian prior on all trainable weights (both encoder and decoder) which are updated to the posterior using the training dataset. For the Bayesian network with parameters $\boldsymbol{\theta}$, posterior $p(\boldsymbol{\theta}~\mid D, x)$, training data with ground truth segmentation masks \(D = (X, Y)\), and sample $x_i$, the predictive posterior distribution for a given ground truth segmentation mask $y_i$ is as follows:
%
\begin{equation}
    p(y_i \mid D, x_i) = \int_{}^{} p(y_i \mid \boldsymbol{\theta}, x_i) p(\boldsymbol{\theta} \mid D) d\boldsymbol{\theta}
\label{eq:posterior}
\end{equation}
While Monte-Carlo dropout~\cite{gal2016dropout} at test-time is a popular approach to approximating this intractable integral, we choose stochastic gradient Monte-Carlo sampling MCMC (SG-MCMC~\cite{welling2011bayesian}) for a better posterior. Stochastic gradient over mini-batches includes a noise term approximating the gradient over the whole training distribution. Furthermore, the cyclical learning rate schedule introduced in~\cite{zhang2019cyclical} known as cyclical SG-MCMC, or cSG-MCMC, allows for faster convergence and better exploration of the multimodal distributions prevalent in deep neural networks. Larger learning step phases provide a warm restart to the subsequent smaller steps in the sampling phases.

The final estimated posterior of the Bayesian network, \(\boldsymbol{\Theta} = \{\boldsymbol{\theta}_1, ... \boldsymbol{\theta}_M\}\), consists of \(M\) moments sampled from the posterior taken during the sampling phases of each learning cycle. With functional model \(\boldsymbol{\Phi}\) representing the neural network, the approximate predictive mean $\mu_i$ for one sample \(x_i\) is:
\begin{equation}
    \mu_i \approx \frac{1}{M} \sum_{m=1}^{M} \boldsymbol{\Phi}_{\theta_m}(x_i)
    \label{eq:predictive_mean}
\end{equation}
% \begin{equation}
%     \hat{y}_i = \textup{argmax}\left ( \mu_i \right )
%     \label{eq:argmax_pred_mean}
% \end{equation}
%
We can derive a segmentation prediction mask $\hat{y}_i$ from $\mu_i$ by taking the maximum output between the foreground and background channels. The predictive uncertainty mask corresponding to this prediction (Equation~\ref{eq:sigma}) represents the \textit{model uncertainty} for the predicted segmentation mask, the variance in the predictive distribution for that sample. 
% TODO insert figure here showing input image, unc mask, and ground truth (to the side) for computing loss and loss contributions.
\begin{figure}[!t]%
\begin{center}
\includegraphics[width=1.0\linewidth]{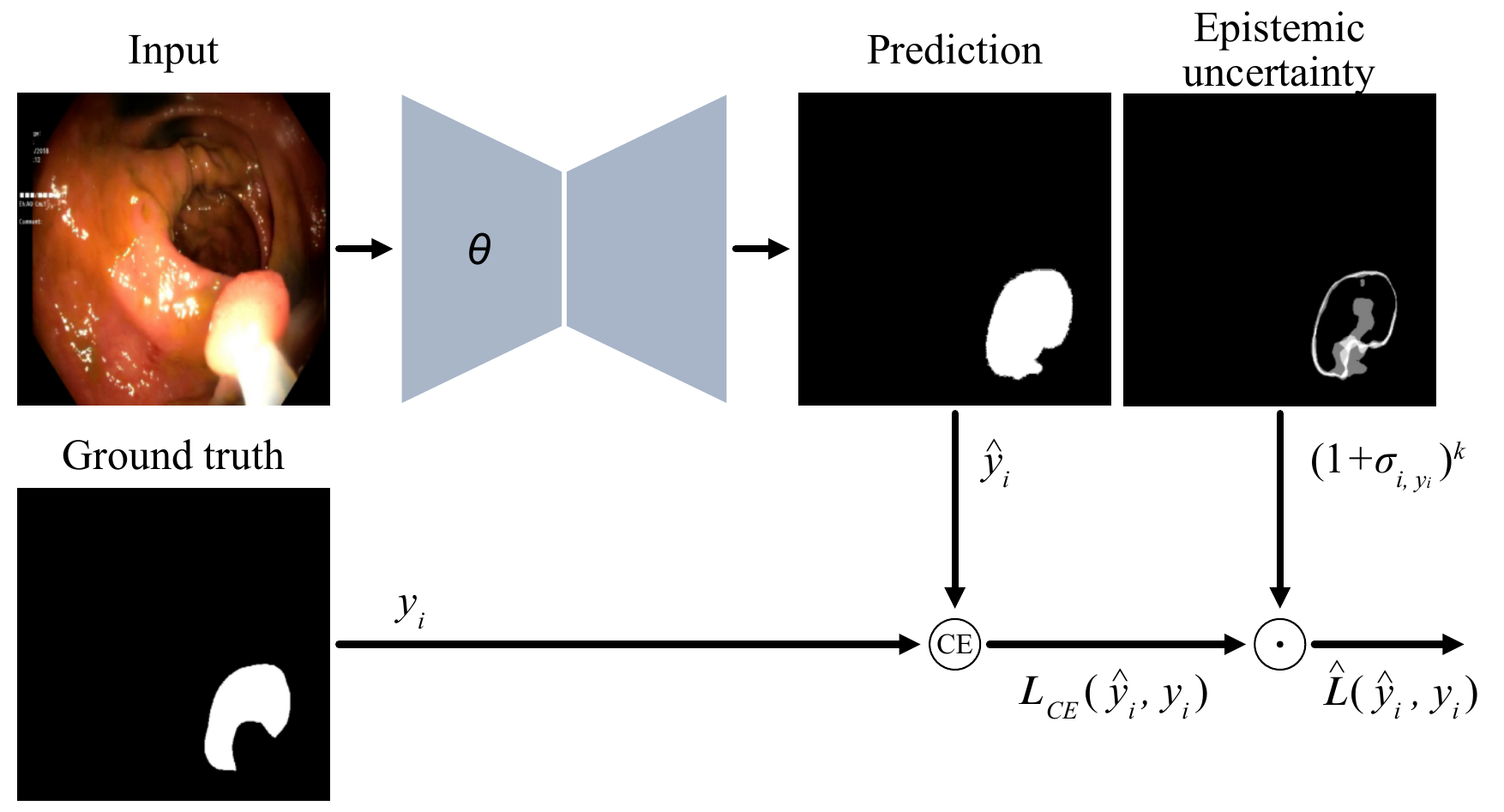} 
\end{center}
\caption{Pixel-wise weighting of cross entropy (CE) loss contribution based on predictive uncertainty maps for each training sample; the model is encouraged to focus on regions for which it is more uncertain.}
\label{fig:units}%
\end{figure}

\begin{equation}
    \sigma_i \approx \sqrt{\frac{1}{M} \sum_{m=1}^{M} \left(\boldsymbol{\Phi}_{\theta_m}(x_i) - \mu_i \right)^2}
\label{eq:sigma}
\end{equation}

We add predictive uncertainty-weighted sample loss~\cite{stone2022epistemic} that identifies high-uncertainty sample regions during training. It also scales the pixel-wise contribution of these regions to the loss computation via a simple weighting function (Equation~\ref{eq:weighted_loss}). This unreduced cross-entropy loss is then averaged over each image and batch (see Fig.~\ref{fig:units}).

\begin{equation}
    \hat{L}(\hat{y}_i, y_i) = L_{CE}(\hat{y}_i, y_i)*(1.0 + \sigma_{i, y_i})^\kappa 
\label{eq:weighted_loss}
\end{equation}

The shift by a constant (1.0) normalises the values, ensuring that the lowest uncertainty samples are never irrelevant to the loss term. $\kappa$ is a tunable de-biasing parameter; \(\kappa = 1\) being a normal weighting, whereas \(\kappa \to \infty\) increases the importance of high-uncertainty regions. As too large a $\kappa$ results in degraded performance due to overfitting, the optimal value is determined by validation metrics.
% 
% TODO move the training test images to supplemental material due to space.
% TODO insert figure showing diversity present in test datasets
\begin{figure}[tb!]%
\begin{center}
\includegraphics[width=1.0\linewidth]{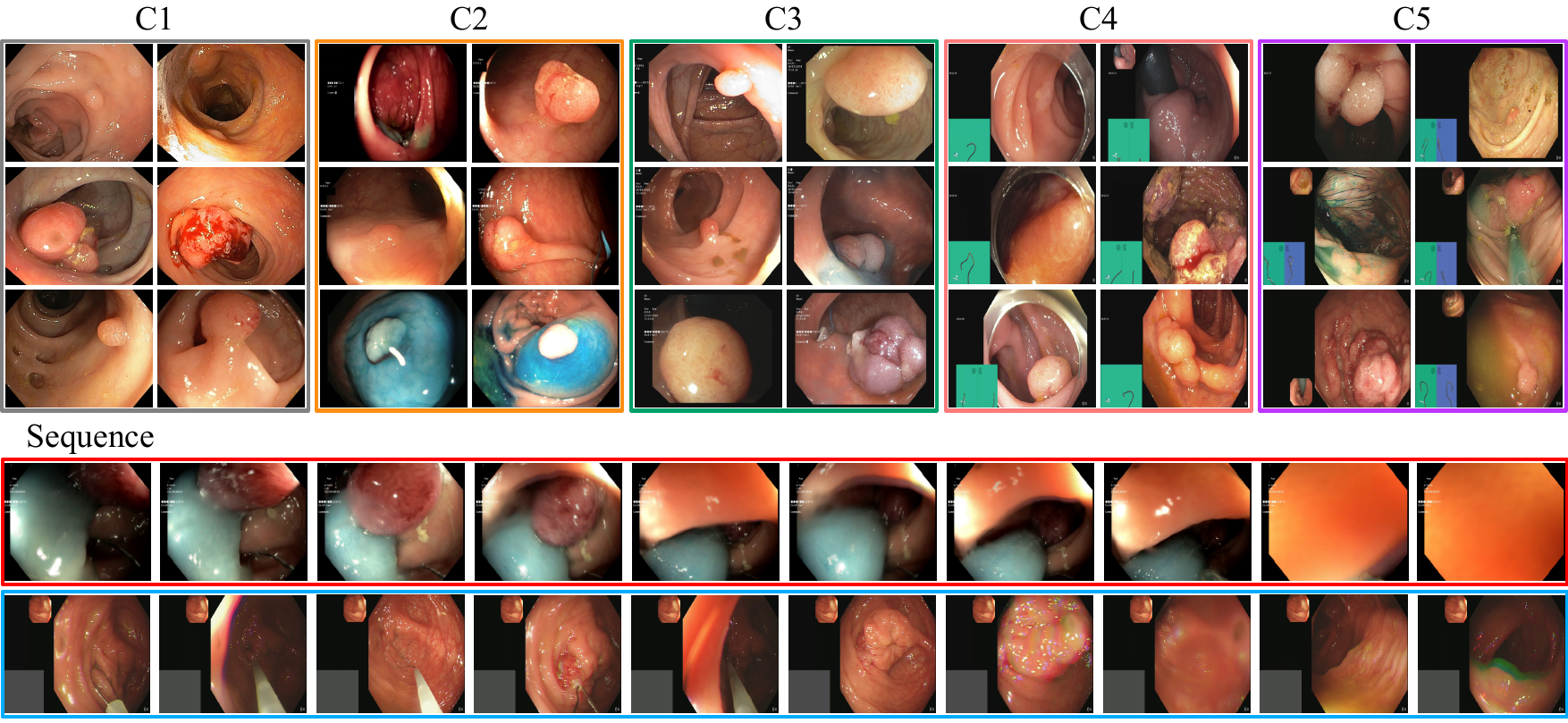}\end{center}
\vspace{-5mm}
%\caption{Samples from the three test sets; from (\textit{top}) C6-SIN, (\textit{middle}) C6-SEQ, and (\textit{bottom}) C1-5-SEQ.}
\caption{Samples from the EndoCV2021 dataset; from (\textit{top}) C1-5 single frames and (\textit{bottom}) C1-5-SEQ; (\textit{top}) highlights the data distribution of each center (C1-C5), which consists of curated frames with well-defined polyps; (\textit{bottom}) demonstrates the variability of sequential data due to the presence of artifacts, occlusions, and polyps with different morphology. }
\label{fig:test_datasets}%
\end{figure}
\section{Experiments and results}
\subsection{Dataset and experimental setup}
PolypGen~\cite{ali2023multi} is an expert-curated polyp segmentation dataset comprising of both single frames and sequence frames (frames sampled at every 10 frames from video) from over 300 unique patients across six different medical centers. 
% The natural data collection format is sequence, and single frames are hand-selected from these to form cleaner, better-quality single frame datasets. 
The natural data collection format is video from which single frames and sequence data are hand-selected.
% Single frames form a cleaner, better-quality with existence of polyp in each sample whereas sequence data comprises of both polyp and non-polyp. 
%
The single frames are clearer, better quality, and with polyps in each frame including polyps of various sizes (10k to 40k pixels), and also potentially containing additional artifacts such as light reflections, blue dye, partial view of instruments, and anatomies such as colon linings and mucosa covered with stool, and air bubbles (Fig.~\ref{fig:test_datasets}). The sequence frames are more challenging and contain more negative samples without a polyp and more severe artifacts, which are a natural occurrence in colonoscopy. Our training set includes 1449 single frames from five centers (C1 to C5) and we evaluate on the three tests sets used for generalisability assessment in literature~\cite{ali2022,ali2023multi}.

The first test dataset has 88 single frames from an unseen center C6 (C6-SIN), and the second has 432 frames from sequence data also from unseen center C6 (C6-SEQ). Here, the first test data (C6-SIN) comprises of hand selected images from the colonoscopy videos while the second test data (C6-SEQ) includes short sequences (every 10$^{th}$ frame of video) mimicking the natural occurrence of the procedure.  The third test dataset includes 124 frames but from seen centers C1 - C5; however, these are more challenging as they contain both positive and negative samples with different levels of corruption that are not as present in the curated single frame training set. 
% which  of the more challenging sequence  but from C1 - C5, centers from which the model has seen single frames during training. 
% 
\begin{table}[ht!]
\begin{adjustbox}{max width=\textwidth}
\begin{tabular}{llcccccccc}
\hline \hline
\multicolumn{1}{c}{\textbf{Dataset}}  &  & \multicolumn{1}{c}{\textbf{Method}}  & & \textbf{JAC} & \textbf{Dice} & \textbf{F2} & \textbf{PPV} & \textbf{Recall} & \textbf{Accuracy} \\ \hline
\multirow{3}{*}{C6-SIN}
&  & SOTA&  & 0.738$\pm$0.3 & 0.806$\pm$0.3 & 0.795$\pm$0.3 & \textbf{0.912$\pm$0.2} & 0.793$\pm$0.3 & \underline{0.979$\pm$0.1}   \\                               &  & BayDeepLabV3+ & & 0.721$\pm$0.3 & 0.790$\pm$0.3  & \textbf{0.809$\pm$0.3} & 0.836$\pm$0.2 & \textbf{0.843$\pm$0.3} & \textbf{0.977$\pm$0.1} \\ 
&      & Ours &  & \textbf{0.740$\pm$0.3} & \textbf{0.810$\pm$0.3} & \underline{0.804$\pm$0.3} & \underline{0.903$\pm$0.1} & \underline{0.806$\pm$0.3} & \textbf{0.977$\pm$0.1} \\ \hline
\multirow{3}{*}{\begin{tabular}[l]{@{}l@{}}C1-5-SEQ\end{tabular}} 
&  & SOTA &  & \textbf{0.747$\pm$0.3} & \textbf{0.819$\pm$0.3} & \textbf{0.828$\pm$0.3} & \underline{0.877$\pm$0.2} & \underline{0.852$\pm$0.3} & 0.960$\pm$0.0 \\
&  & BayDeepLabV3+ &  & 0.708$\pm$0.3 & 0.778$\pm$0.3 & 0.805$\pm$0.3 & 0.784$\pm$0.3 & \textbf{0.885$\pm$0.2} & \textbf{0.963$\pm$0.0} \\
&  & Ours  &  & \underline{0.741$\pm$0.3} & \underline{0.810$\pm$0.3} & \underline{0.815$\pm$0.3} & \textbf{0.888$\pm$0.2} & 0.836$\pm$0.3 & \underline{0.961$\pm$0.0} \\ \hline 
\multirow{3}{*}{C6-SEQ} 
&  & SOTA &  & 0.608$\pm$0.4 & 0.676$\pm$0.4 & 0.653$\pm$0.4 & \underline{0.845$\pm$0.3} & 0.719$\pm$0.3 & 0.964$\pm$0.1 \\ 
&  &  BayDeepLabV3+ &  & \underline{0.622$\pm$0.4} & \underline{0.682$\pm$0.4} & \underline{0.669$\pm$0.4} & 0.802$\pm$0.3 & \textbf{0.764$\pm$0.3} & \underline{0.965$\pm$0.1} \\ 
&  & Ours & & \textbf{0.637$\pm$0.4} & \textbf{0.697$\pm$0.4} & \textbf{0.682$\pm$0.4} & \textbf{0.858$\pm$0.3} & \underline{0.741$\pm$0.3} & \textbf{0.967$\pm$0.1} \\\hline\hline\\
\end{tabular}
\end{adjustbox}
\caption{Evaluation of the state-of-the-art deterministic DeepLabV3+, BayDeepLabV3+, and our proposed BayDeepLabV3+Unc, showing mean and standard deviations across the respective test dataset samples. \textbf{First} and \underline{second} best results for each metric per dataset formatted.}
\label{tab:results}
\end{table}
As no C6 samples nor sequence data are present in the training data, these test sets present a challenging generalisability problem.~\footnote{C1-5-SEQ and C6-SEQ data are referred to as DATA3 and DATA4, respectively, in~\cite{ali2022}}.

Training was carried out on several IBM Power 9 dual-CPU nodes with 4 NVIDIA V100 GPUs. Validation metrics were used to determine optimal models for all experiments with hyper-parameters chosen via grid search. Perhaps due to some frames containing very large polyps with high uncertainties, we found that the gradients of Bayesian models with uncertainty-weighted loss (BayDeepLabV3+Unc) occasionally exploded during the second learning cycle, and clipping the absolute gradients at 1.0 for all weights prevented this issue. All Bayesian DeepLabV3+ (BayDeepLabV3+) models had 2 cycles, a cycle length of 550 epochs, noise control parameter $\alpha$ = 0.9, and an initial learning rate of 0.1. For BayDeepLabV3+Unc, we found optimal results with de-biasing tuning parameter $\kappa$ = 3. Posterior estimates for BayDeepLabV3+ and BayDeepLabV3+Unc included 6 and 4 samples per cycle, respectively.
\begin{figure}[h!]%
\begin{center}
% \begin{subfigure}[b]{0.49\textwidth}
%          \centering
%          \includegraphics[width=\textwidth]{imgs/C6vsDATA4.pdf}
%          \caption{Comparison between single vs sequence modalities}
%          \label{fig:single_vs_seq}
%      \end{subfigure}
%      \hfill
%      \begin{subfigure}[b]{0.49\textwidth}
%          \centering
%          \includegraphics[width=\textwidth]{imgs/DATA3vsDATA4.pdf}
%          \caption{Comparison between in vs out of distribution}
%          \label{fig:in_vs_ood}
%      \end{subfigure}
\includegraphics[width=\textwidth]{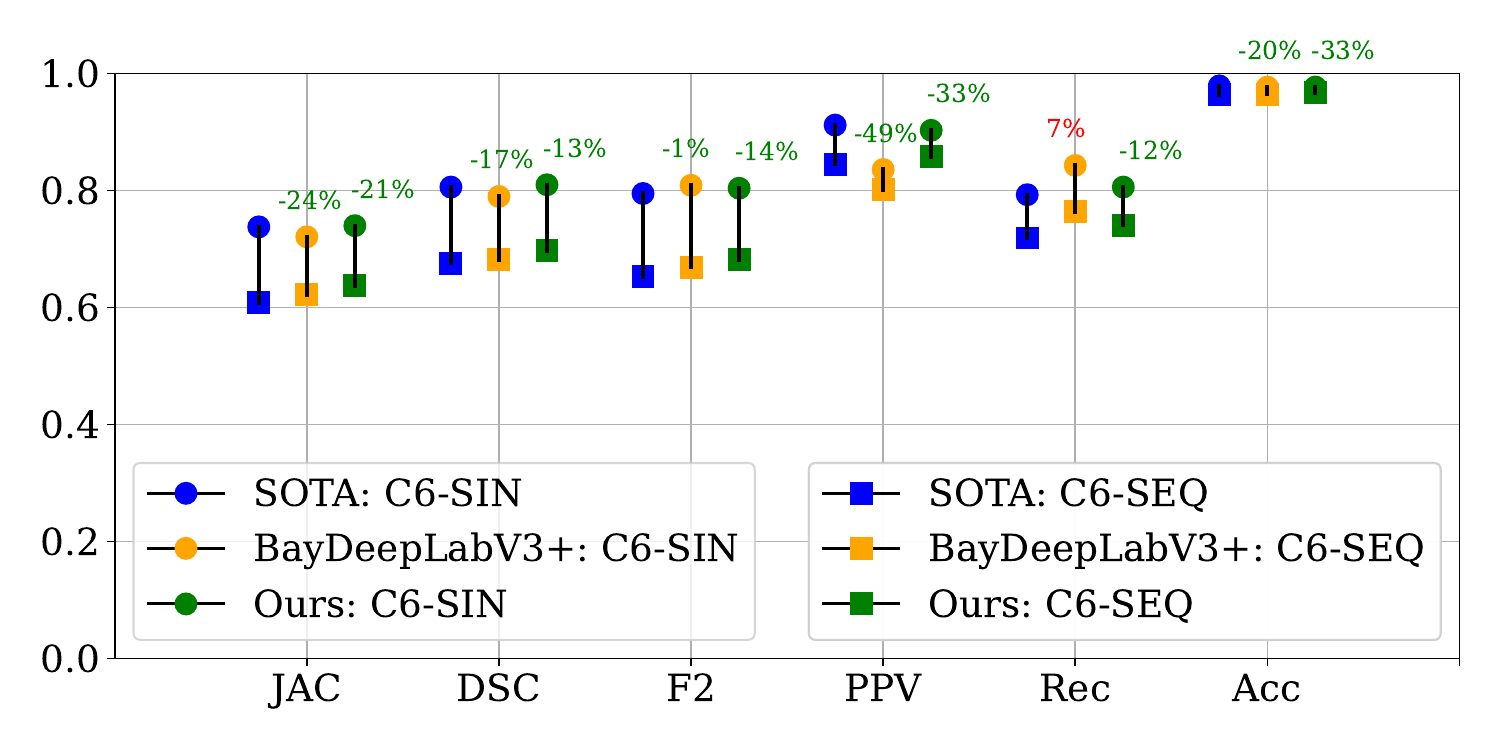}
\includegraphics[width=\textwidth]{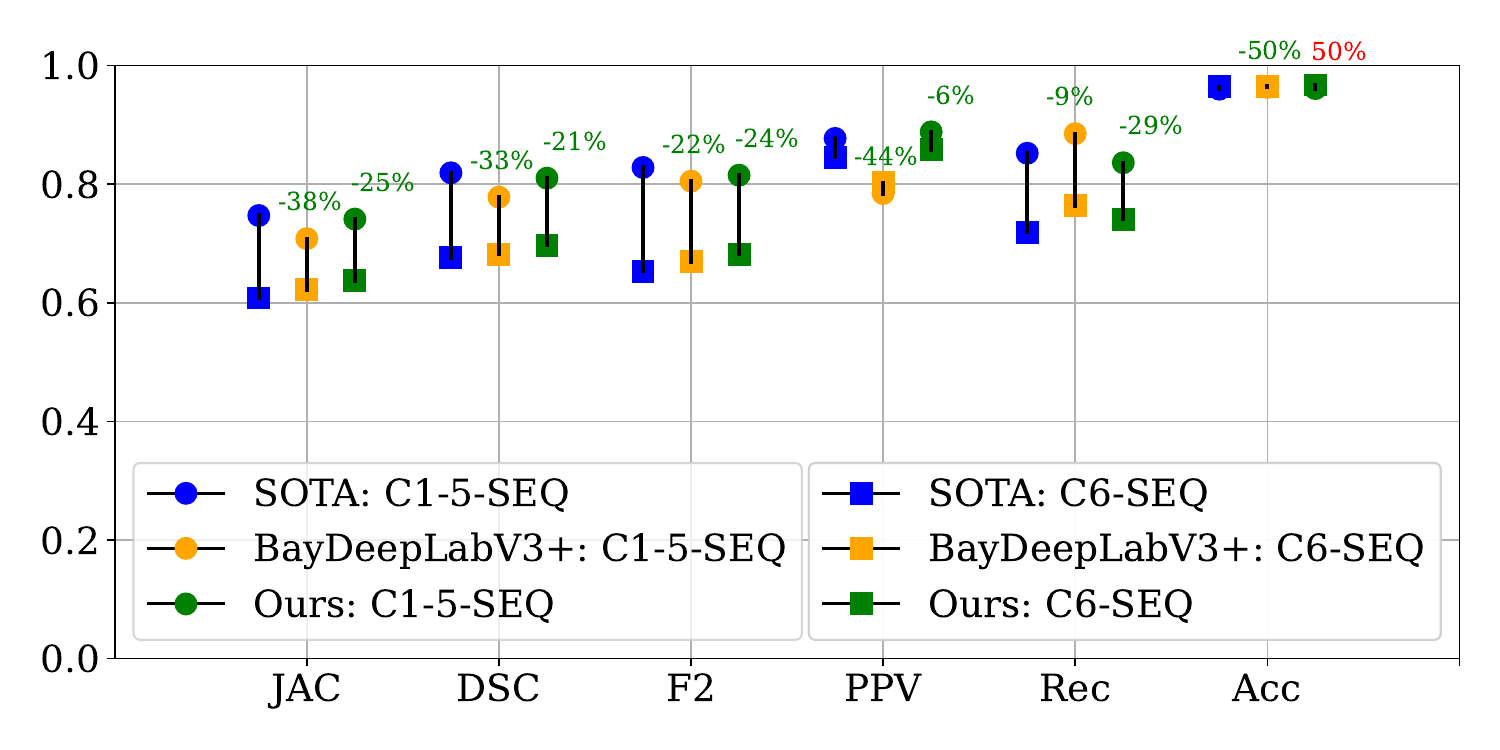}
\end{center}
\vspace{-5mm}
\caption{Performance gaps of the three models (state-of-the-art deterministic DeepLabV3+, BayDeepLabV3+, and BayDeepLabV3+Unc) between the three different test sets; \textit{(top)} comparing performance on single vs. sequence frames from out-of-distribution test set C6 (C6-SIN vs. C6-SEQ), and \textit{(bottom)} sequence frames from C1 - C5 vs. unseen C6 (C1-5-SEQ vs. C6-SEQ). The subtext above bars indicates the percent decrease in performance gap compared to SOTA; a larger percent decrease and shorter vertical bar length indicate better generalisability.}
\label{fig:gaps}%
\end{figure}
\subsection{Results}
% \subsection{Comparison with SOTA methods}
We use the state-of-the-art deterministic model~\footnote{https://github.com/sharib-vision/PolypGen-Benchmark} and checkpoints to evaluate on the three test sets, and compare against the baseline Bayesian model BayDeepLabV3+ and BayDeepLabV3+Unc with uncertainty-weighted loss. 
% ~\footnote{https://github.com/sharib-vision/PolypGen-Benchmark}

We report results for Jaccard index (JAC), Dice coefficient (Dice), F$_\beta$-measure with $\beta$ = 2 (F2), positive predictive value (PPV), recall (Rec), and mean pixel-wise accuracy (Acc). PPV in particular has high clinical value as it indicates a more accurate delineation for the detected polyps. Recall and mean accuracy are less indicative since the majority of frames are background in the segmentation task and these metrics do not account for false positives. A larger number of false positive predictions can cause inconvenience to endoscopists during colonoscopic procedure and hence can hinder clinical adoption of methods.
% False positives correspond to a detected area larger than the actual polyp and are not appreciated in clinical adoption.
% 
Figure~\ref{fig:gaps} illustrates that our approach maintains SOTA performance across most metrics and various test settings, even outperforming in some cases; simultaneously, the performance gaps between different test sets representing different challenging features (1) image modalities (single vs. sequence frames) and (2) source centers (C1 - C5 vs. C6) are significantly decreased. Simply turning the SOTA model Bayesian improves the model's ability to generalise, yet comes with a sacrifice in performance across metrics and datasets. Our proposed uncertainty-weighted loss achieves better generalisability without sacrificing performance (also see Table~\ref{tab:results}). We note performance superiority to SOTA especially on C6-SEQ, approximately 3\% improvement on Dice. We can also observe slight improvement on PPV for test sets with sequence (both held-out data and unseen centre data). Finally, we note that in clinical applications, the uncertainty maps for samples during inference could be useful for drawing clinicians' attention towards potentially challenging cases, increasing the likelihood of a fairer outcome.

\section{Conclusion}
We have motivated the critical problem of model fairness in polyp segmentation on a multi-center dataset, and modified a Bayesian bias mitigation method to our task. The results on three challenging test sets show strong potential for improving generalisability while maintaining competitive performance across all metrics. Furthermore, the proposed mitigation method is implicit, not requiring comprehensive knowledge of biases or out-of-distribution features in the training data. This is of particular importance in the medical community given the sensitivity and privacy issues limiting collection of annotations and metadata. Our findings are highly relevant to the understudied problem of generalisation across high variability colonoscopy images, and we anticipate future work will include comparisons with other methods to improve generalisability and an extension to the approach. We also anticipate having access to additional test data for more in-depth analysis of the results.

\subsubsection*{Acknowledgements}
R. S. Stone is supported by an Ezra Rabin scholarship.

\bibliographystyle{splncs04}
\bibliography{ref}
\end{document}